\documentclass[nohyperref]{article}

\usepackage{microtype}
\usepackage{graphicx}
\usepackage{subfigure}
\usepackage{booktabs} 

\usepackage{hyperref}

\usepackage[accepted]{icml2022}

\usepackage{amsmath}
\usepackage{amssymb}
\usepackage{mathtools}
\usepackage{amsthm}
\usepackage{tabularx}
\usepackage{multirow}
\newcommand{\argmin}[1]{\underset{#1}{\operatorname{arg}\,\operatorname{min}}\;}

\DeclarePairedDelimiterX{\norm}[1]{\lVert}{\rVert}{#1}

\usepackage[capitalize,noabbrev]{cleveref}

\theoremstyle{plain}

\theoremstyle{definition}

\theoremstyle{remark}

\usepackage[textsize=tiny]{todonotes}

\icmltitlerunning{KENN: Enhancing Deep Neural Networks by Leveraging Knowledge for Time Series Forecasting}

\begin{document}

\twocolumn[
\icmltitle{KENN: Enhancing Deep Neural Networks by Leveraging Knowledge for Time Series Forecasting }



\icmlsetsymbol{equal}{*}

\begin{icmlauthorlist}
\icmlauthor{Muhammad Ali Chattha}{nust,tukl,dfki}
\icmlauthor{Ludger van Elst}{dfki}
\icmlauthor{Muhammad Imran Malik}{nust}
\icmlauthor{Andreas Dengel}{tukl,dfki}
\icmlauthor{Sheraz Ahmed}{dfki}

\end{icmlauthorlist}

\icmlaffiliation{nust}{National University of Science and Technology, NUST, Islamabad, Pakistan}
\icmlaffiliation{tukl}{Technical Universität Kaiserslautern, TUKL, Kaiserslautern, Germany}
\icmlaffiliation{dfki}{Deutsches Forschungszentrum für Künstliche Intelligenz, DFKI, Kaiserslautern, Germany}

\icmlcorrespondingauthor{Muhammad Ali Chattha}{muhammad\_ali.chattha@dfki.de}

\icmlkeywords{Machine Learning, ICML}

\vskip 0.3in
]



\printAffiliationsAndNotice{}  

\begin{abstract}
End-to-end data-driven machine learning methods often have exuberant requirements in terms of quality and quantity of training data which are often impractical to fulfill in real-world applications. This is specifically true in time series domain where problems like disaster prediction, anomaly detection, and demand prediction often do not have a large amount of historical data. Moreover, relying purely on past examples for training can be sub-optimal since in doing so we ignore one very important domain i.e knowledge, which has its own distinct advantages. In this paper, we propose a novel knowledge fusion architecture, Knowledge Enhanced Neural Network (KENN), for time series forecasting that specifically aims towards combining strengths of both knowledge and data domains while mitigating their individual weaknesses. We show that KENN not only reduces data dependency of the overall framework but also improves performance by producing predictions that are better than the ones produced by purely knowledge and data driven domains. We also compare KENN with state-of-the-art forecasting methods and show that predictions produced by KENN are significantly better even when trained on only 50\% of the data.
\end{abstract}

\section{Introduction}
\label{sec:intro}

Time series forecasting has remained an important area of research as it directly deals with problems such as demand prediction~\cite{lu2021hybrid}, resource optimization~\cite{xiang2021urban}, traffic flow predictions~\cite{zheng2021joint}, predictive maintenance~\cite{arena2022novel}, etc which are critical for businesses, governments, and industries. Having an accurate estimate of future trends can help mitigate losses, increase financial profits, and can allow for effective planning and resource utilization. As a result, it is not surprising that a great deal of emphasis has been laid on time series forecasting problems and improvements to forecasting methods are highly desirable.

Recently, purely data driven Machine Learning(ML) methods like Deep Neural Networks (DNNs) have gained huge success. DNNs are quite proficient in extracting useful features from the data and have achieved amazing performance in various domains that include time series forecasting problems~\cite{cai2020traffic,yu2017spatio}, image segmentation and classification~\cite{yuan2021segmentation,dai2021coatnet}, natural language processing~\cite{du2021all}, etc. However, more often than not these networks are huge, containing millions of trainable parameters whose optimization requires an equally large training dataset which in many real-world applications are not available. Arguably, dependency on a large amount of accurately labeled data is one of the biggest limitations of DNNs. Dependency on large datasets puts DNNs in quite a predicament for time series modeling problems since a long time series in temporal domain may still have very few data points for DNN to train upon. For example, a monthly time series spanning over 20 years will only have 240 historical observations for training. Consequently, complex deep networks are prone to overfitting on temporal forecasting problems and in many real-world forecasting problems their superiority, if at all, is not as profound as in other domains\cite{makridakis2000m3,makridakis2018m4}. 

On the other hand, Knowledge Driven Systems(KDS) aim to rely on human knowledge to come up with predictions. They do not normally rely on historical data. KDS typically comprises of a knowledge base that consists of problem-specific facts and manually defined rules. These rules are specifically tailored towards capturing the knowledge of human experts and are followed in a predefined manner for inference in an attempt to mimic the decision making process of the experts. KDS systems are still widely used especially in risk critical domains such as health~\cite{zhang2019common,fernandes2018c}, collision avoidance systems~\cite{hagen2018mpc}, etc.  For time series forecasting, KDS can have knowledge in the form of If-Then conditioning statements or arithmetic functions drawn from the statistical corpus. Although these systems do not directly or marginally rely on data, however, formulating rules that generalize for every scenario is an arduous task. Needless to say, both knowledge and data driven domains have their distinct advantages that are complementary in nature. DNNs are dexterous in processing and extracting useful features from the data while KDS can model the underlying process very well which enables them to work well in data scarce scenarios. Hence relying on one domain to come up with a solution can be sub-optimal.

A natural step forward is to combine DNNs and KDS. Hybrid schemes that combine additional information with DNNs are becoming increasingly common. However, most of these hybrid schemes for time series forecasting rely on ensemble methods, where separate systems are combined post predictions by using weighted ensemble~\cite{kaushik2020ai, choi2018combining, smyl2020hybrid}, or by using statistical methods for preprocessing and feature extraction which are then given as inputs to DNNs~\cite{tripathy2018use,smyl2020hybrid}. Although these frameworks combine different domains, they are restricted by the information present in the data. Cases where data is limited severely hamper their performance. Moreover, we believe that in an ideal fusion scheme one domain should be aware of the weaknesses of the other and should try to supplement the missing information. Perhaps close to our objective are techniques revolving around knowledge distillation~\cite{hinton2015distilling} which work in a teacher-student setting. Here student network tries to mimic predictions made by the teacher network, that may be based on KDS or a more complex deep network. We argue that although knowledge distillation improves the capabilities of student network it, however, intrinsically assumes that the teacher network is accurate. Cases, where predictions made by teacher networks are inaccurate or unreliable, are not catered. Such scenarios are not uncommon in KDS since it is hard to formulate rules that capture all of the knowledge.

Keeping in mind the limitations of current fusion schemes and the importance of the forecasting problem, we propose Knowledge Enhanced Neural Network(KENN) a novel framework for fusing knowledge from KDS into DNNs for time series forecasting. KENN specifically aims towards mitigating data dependency of DNNs without introducing an additional dependency on the accuracy of the KDS. KENN achieves this by incorporating predictions given by KDS in a residual scheme that allows it to assess the efficacy of KDS and fill the information missing from the knowledge domain with the one present in the data. We show that KENN is capable of combining strengths of both knowledge and data domains while suppressing their individual weaknesses. KENN reduces data dependency of DNN while also catering to scenarios where knowledge in KDS is limited or unreliable. Although this research aimed to come up with a fusion scheme that attempts to combine the best of both worlds rather than proposing the best DNN or KDS model for forecasting, however, we still compare KENN with other State-Of-The-Art (SOTA) methods on a range of different datasets and demonstrate its effectiveness. In particular, the contributions of this paper are as follows:
\begin{itemize}

\item We introduce a novel knowledge fusion framework, Knowledge Enhanced Neural Network (KENN), that allows for integration of knowledge with DNNs for time series forecasting application where individual strengths of both domains are retained.
\item we show that KENN reduces the dependency of DNN on historical data while simultaneously reducing accuracy constraints on KDS. In other words, we show that KENN tries to cater for shortcomings of both knowledge and data driven methods.
\item We show that KENN is fully flexible and dynamic in terms of integrating different sources of KDS and DNN architectures. 
\item We compare KENN with the SOTA models on various real-world time series forecasting datasets and show that in most settings KENN beats SOTA by a significant margin even when trained on a reduced training set.
\end{itemize}

\section{Related work}\label{sec:relatedwork}

In this section, we review some of the relevant work in the literature specifically pertaining to knowledge sharing including multi-modal learning approaches. 

\textbf{Knowledge Distillation} based techniques aim to transfer knowledge from the teacher network to a student network. This knowledge is typically transferred in a process where teacher produces soft predictions that the student network tries to emulate.~\cite{hu2016harnessing} utilized iterative knowledge distillation to transfer knowledge. They used a teacher network that was made from first-order logic rules and used it to train the student DNN model. They updated both the teacher as well as the student at each iteration during the learning process. The main attempt was to find a teacher network that can match the rule set in terms of predictions while not diverging significantly from the labels in the data. Similarly, \cite{xie2019self} used a teacher network that was trained on JFT-300m dataset that contained 300 million images. After which they used a teacher student setting on ImageNet dataset where the teacher network produced soft predictions that student network followed. Knowledge distillation can also be in the form of feature map transfer that provides student network an attention map for intermediary layers. As a result the student network learns which part of the intermediate feature map does the teacher network pay attention too. Such approach is used in the work presented by \cite{chen2021cross}. However, mostly distillation method rely on minimizing KL-divergence between the distributions of the prediction made by the teacher and the student to make the two distributions similar. KL-divergence can not be directly applied for forecasting task since the output of the network is not a distribution but instead is an estimate of future time series values. Moreover, Since the task of the student network in this framework is to emulate the predictions made by the teacher, this leads to the strength of the student network being ignored by the system and scenarios where teacher is flawed or unreliable is not catered in distillation setting.

\textbf{Attention based models} have gained considerable popularity recently.~\cite{li2019enhancing} proposes a convolutional self-attention mechanism for transformer networks that used causal convolutions of different kernel sizes. Using this self-attention mechanism the proposed transformer network was able to achieve better performance on forecasting datasets. Similarly, work proposed by ~\cite{zheng2020gman,li2019forecaster,cai2020traffic} also made use of transformers networks specifically for vehicular traffic forecasts. The main focus in these works were to capture the spatial dependencies along with temporal ones to model the traffic flow system. These methods used graph based convolutional networks to model the spatial dependency with ~\cite{zheng2020gman} using both spatial and temporal attention mechanism, while ~\cite{li2019forecaster} and ~\cite{cai2020traffic} utilizing temporal attention. They claimed SOTA results on different traffic forecasting datasets. Highlighting parts in the inputs by utilizing spatial attention or self temporal attention is effective in making the network focus on important portions of the input which in turn results in better forecasts. However, since these techniques only rely on the historical data they do not cater for scenarios where data is scarce and may not contain enough information for DNNs to retrieve. Moreover, using spatial information makes sense in traffic datasets since roads in a certain area are interconnected and hence traffic information on certain road is helpful when predicting traffic flow on the other connected road, however, this scheme will not work if there is no mutual dependence among time series in a dataset. 

\textbf{Knowledge based nerual architectures} incorporate knowledge in the form of logic rules directly into neural network architectures. \cite{towell1994knowledge} proposed the Knowledge-based Artificial Neural Networks (KBANN) architecture where logic rules were structured hierarchically. Multi Layer Perceptron (MLP) based network is then constructed where its layers match the hierarchy of rule set. Each of the element and their relation in the rule set is represented by a neuron and its weight, respectively, in the MLP architecture. To allow MLP model to learn some features from the data as well, additional neurons are also introduced whose weights were learned during the training phase. Similarly, ~\cite{tran2018deep} constructs neural network with similar concept. Here too, a set of logic rules is employed in conjunction with the neural network. Although, these architectures directly incorporate knowledge into the architecture of the neural network however, having neural network that corresponds in terms of layers and architecture limits applicability of such techniques. This introduces constraints on neural networks architecture as it has to be strictly defined by the rule set. Moreover, it also limits use of a different knowledge base that may not necessarily have a hierarchical layout.   


\textbf{Ensemble} based methods combine predictions of different models. Normally they rely on taking a weighted average of the predictions made by the models employed \cite{larrea2021extreme,kaushik2020ai, smyl2020hybrid}. Although, ensemble methods have gained a lot of traction recently due to their ability to improve the results, we believe that in ideal scenario one network should be aware of the strengths of the other model and try to focus on the information missing in the other domain. Moreover, if one of the model in the ensemble is bad in terms of prediction it can degrade the overall performance of the ensemble rather than improving. We believe that fusion scheme should be intelligent enough to assess the efficacy of the constituent models and incorporate information accordingly.

\section{Problem Formalization}\label{sec:ProblemFormalization}

In this section, we mathematically formalize the forecasting problem. Let $\mathcal{X}$ be a list containing ${x_{t}, x_{t-1}, ..., x_{t-w}}$ time series observations. Here $x_{t}$ represent the time series observation at time $t$ and $w$ represents the total number of past observations used for prediction. The aim of regression system is to learn a transfer function that maps $\mathcal{X}$ to $\hat{\mathcal{Y}}$, where $\hat{\mathcal{Y}}$ represents a list containing predicted values $\hat{x}_{t+1},\hat{x}_{t+2},...,\hat{x}_{t+h}$. $h$ represents horizon i.e number of future observations being predicted. Mathematically, this mapping function can be represented as

\begin{equation}
    [\hat{x}_{t+1},\hat{x}_{t+2},...,\hat{x}_{t+h}]=\Phi([x_{t}, x_{t-1}, ..., x_{t-p}]; \mathcal{W})
\end{equation}

\begin{equation}
     \hat{\mathcal{Y}} = \Phi(\mathcal{X}; \mathcal{W})
\end{equation}
 
\noindent where $\mathcal{W} = {\{W_l, b_l\}}_{l=1}^{L}$ represents trainable weights of the neural network comprised of L layers and $\Phi:\mathbb{R}^{w+1}\mapsto \mathbb{R}^{h}$ defines the mapping from the input space to the output space. The objective of the regression system is to learn optimal parameters $\mathcal{W}^{*}$ that makes the predictions $\hat{\mathcal{Y}}$ as close as possible to $\boldsymbol{\mathcal{Y}}$, the ground truth. These optimal parameters $\mathcal{W}^{*}$ are computed in a recursive manner using gradient descent where the prediction loss is calculated after every iteration and the parameters $\mathcal{W}$ are adjusted to minimize the loss. This process is repeated until the loss plateaus. Typically, Mean Squared Error (MSE) is used as a loss function, and hence, the optimization problem can be stated as:

\begin{equation}\label{unconditionedOptimProblem}
      \mathcal{W}^{*} = \argmin{\mathcal{W}} \sum_{x \in \mathcal{Y}}\norm{\boldsymbol{\mathcal{Y}} - \Phi([x_{t},..., x_{t-p}]; \mathcal{W})}_2^2
\end{equation}

\begin{equation}\label{eq:simple_opt}
     \mathcal{W}^{*} = \argmin{\mathcal{W}} \sum_{x \in \mathcal{Y}} \norm{\boldsymbol{\mathcal{Y}} - \hat{\mathcal{Y}}}_2^2
\end{equation}
 
\noindent where $\boldsymbol{\mathcal{Y}}$ and $\hat{\mathcal{Y}}$ denotes the ground truth and the predicted values respectively. 

\section{KENN: The Proposed Framework}\label{sec:Kenn}

In this section, we discuss at length the constituent data and knowledge driven systems and introduce the knowledge fusion framework, KENN. We first introduce the time series dataset used for evaluating KENN followed by descriptions of KDS and DNN used in our testing. Lastly, we elaborate on the knowledge fusion mechanism employed in KENN. 

\subsection{Dataset}\label{sec:dataset_KENN}

We employed a real-world traffic dataset from Caltrans Performance Measurement System (PeMS). PeMS system monitors the flow of traffic on California Highways by using a series of sensors at different locations. For our evaluation, we used readings of one of the sensors that measure traffic flow on Richards Ave. The whole dataset spanned a time period of nine months and was grouped into 30-minute windows with the forecasting goal of predicting the average number of vehicles for the next 30 minutes. We do our initial evaluations on this dataset. After which we employ three widely used real-world forecasting benchmark datasets and compare our techniques with other SOTA methods. These datasets belonged to three different application domains namely traffic, energy, and stocks and each one contained a multitude of time series. More details on the datasets can be found in the appendix.

\subsection{Knowledge Driven System}\label{sec:KDS}

KDS normally consists of a knowledge base that is crafted by human experts, who have considerable experience and knowledge about the underlying problem. The knowledge base mainly consists of manually defined rules which aim to capture the knowledge of domain experts. The inference is made based on these rules defined in the knowledge base. For time series forecasting, the rules contained in the knowledge base of the KDS can be in the form of conditional If-Then statements~\cite{moghram1989analysis,zhang2017novel}, where knowledge is defined by logical expressions, or it can be in the form of statistical methods~\cite{box2015time,hunter1986exponentially}, where knowledge is defined by arithmetic operations. However, any other form that is capable of knowledge representation can be used in a KDS like ontologies, relational databases, etc. 

For our evaluation, we formulate the knowledge base by using concepts drawn for graph theory and also employ a simple ruleset that we believe models the traffic flow at a very rudimentary level. In graph-based models, the relations between objects are modeled using a graph structure where the objects are represented as nodes and the relation between the objects is represented by edges. The nodes that have some similarities with other nodes are connected through edges. Edges have some weight associated with them, called edge weights, that capture the intensity of the similarity or relation between the nodes. In our case,  we first consider each observed value in the time series as nodes where the height of the node represents the value of the observation. The edges or links between the nodes are calculated by using a partial auto correlation function. Those nodes that correlate greater than a certain threshold are connected. The value of the partial auto correlation function is used as edge weights. An additional distance penalty is also imposed where linked nodes that are far apart from each other in time are penalized and the weight of their connecting edge is reduced. Predictions are made by summing the values of connected nodes multiplied by their respective edge weights. Apart from this additional rules are also defined such as at the beginning of office time the mean of traffic prediction given by graph-based prediction should be close to the mean traffic flow during the same time of the previous day etc. The final predictions of the KDS are given by the graph-based predictions that are conditioned to the rules defined in the ruleset. We provide additional information along with the mathematical formulation of our KDS in the appendix. 

Since we claim KENN to be agnostic to the type of KDS used, we also employ different KDS based on statistical method namely seasonal ARIMA, and evaluate the results in section\ref{sec:Ablation} where we perform ablation studies. It is important to note that we do not claim to be experts in the traffic management system or that the KDS used in this paper is groundbreaking, in fact, we only demonstrate the effectiveness of our knowledge fusion mechanism that can make use of simple information present in KDS, as in our case, to improve the performance of DNNs as shown in the results section \ref{sec:Results}.
\subsection{Deep Neural Network}\label{sec:DNN}

We employ Long Short Term Memory (LSTM) based model as our ML method. We spent considerable effort in finding the optimal network parameters through grid search over reasonable hyperparameter search space. We specifically aimed towards finding optimal figures for the number of layers, the number of units in each layer, activation functions, and the loss metrics. The final LSTM architecture consisted of 3 layers followed by a dense layer that gave the final predictions. The number of units within each LSTM layer was \{512,64,32\} and we used Rectified Linear Activation Unit (ReLU) as the activation function for all the layers. Training samples from the data were obtained using a rolling window approach. Here a window of size $input\_windowSize$ and $Horizon$ were used to get the input sample and the corresponding label respectively. The remaining samples are obtained by sliding both of these windows by an index of 1 until the entire dataset was covered. For normalization, we used Min-Max scaling that was applied to each training sample. 
We also used TCN based architecture as our ML based method in the ablation studies section \ref{sec:Ablation}.

\subsection{KENN: Knowledeg Enhanced Neural Network}\label{sec:KennFusion}

\begin{figure}[ht]
\label{fig:arch}
\vskip 0.2in
\begin{center}
\centerline{\includegraphics[width=\columnwidth]{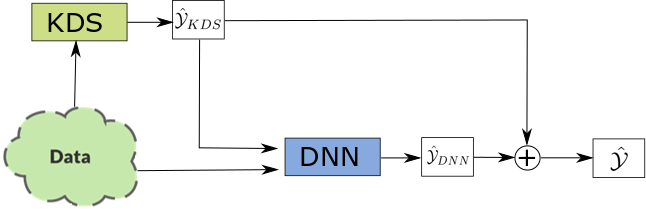}}
\caption{KENN Framework}

\end{center}
\vskip -0.2in
\end{figure}
Fig \ref{fig:arch} shows the architecture of the proposed KENN framework. First, the historical data is shown to KDS which, based on rules defined in its knowledge base, comes up with predictions. These predictions given by KDS are then given to DNN at the input, where it is concatenated with the input training sample, and also added to the out of the DNN resulting in a residual scheme. As a result, the learning objective of the DNN changes, and instead of learning complete input space to output space transfer function the DNN now only learns the residual function. Since the final predictions given by KENN are the summation of predictions given by KDS and DNN, equation \ref{eq:simple_opt} can now be written as
\begin{equation}\label{eq:kenn_opt}
\begin{split}
    \mathcal{W}^{*} = \argmin{\mathcal{W}} \sum_{x \in \mathcal{Y}} \norm{\boldsymbol{\mathcal{Y}} - (\hat{\mathcal{Y}}_{KDS}+\hat{\mathcal{Y}}_{DNN})}_2^2 \\
    = \argmin{\mathcal{W}} \sum_{x \in \mathcal{Y}} \norm{\boldsymbol{(\mathcal{Y}} - \hat{\mathcal{Y}}_{KDS})-\hat{\mathcal{Y}}_{DNN})}_2^2 \\
     = \argmin{\mathcal{W}} \sum_{x \in \mathcal{Y}} \norm{\xi_{KDS}-\hat{\mathcal{Y}}_{DNN})}_2^2
\end{split}
\end{equation}

where $\xi_{KDS}$ represents the error in the predictions given by KDS. As evident from Eq. \ref{eq:kenn_opt} the overall objective of KENN is to minimize the error present in knowledge based predictions instead of learning the complete input to the output transfer function. In other words, the KENN framework allows Knowledge and data driven domains to work in complementary manner, where the data driven domain assesses the weaknesses of the knowledge domain and tries to supplement the shortcomings with its own strengths i.e extracting information from the data. However, not every information present in the historical data is processed instead strengths of DNN are used with a very focused objective which is to only find the information missing from the knowledge domain. KENN promotes strengths of both knowledge and data domains which we believe is essential for reaching a more natural fusion mechanism. Offsetting information missing in one with one present in the other allows KENN to tackle scenarios where information present in KDS is unreliable or where historical data is scarce. In the following section, we do a thorough evaluation of KENN and back the claims with experimental results.

\section{Results}\label{sec:Results}

To evaluate the performance of KENN, we curate a range of different experimental settings each with its unique constraints. These constraints are specially aimed to capture different real-world settings. To highlight the benefits of using KENN, we also report results achieved by the respective knowledge and the data driven network when they are evaluated independently. For a fair comparison, all of the preprocessing and network hyperparameters were kept the same for the DNN model when it was tested in isolation and when employed in KENN. 

In the first setting, we test KENN in a normal scenario where KDS is not handicapped and all of the training data is available for KENN to train upon. To evaluate KENN's performance in data-scarce scenario, we reduce the amount of training data provided to the models for training. We present the findings from this experiment in case 2. After this, we deliberately worsen the performance of KDS by injecting random noise into its predictions to simulate a setting where knowledge in KDS is limited or inaccurate. We present the results under this setting in case 3. A direct extension of the last two experiments is to evaluate KENN performance in the worst case scenario where both of the conditions hold i.e. the amount of training data is scarce and the knowledge in KDS is unreliable. We summarize the results of this experiment in case 4. Finally, we evaluated KENN's performance in cases where the expert contained redundant or no information at all. To achieve these scenarios, we forced KDS to repeat the last observed value in the time series as its prediction essentially producing redundant information. In the next experimental setting, we forced KDS to always predict zero values simulating a scenario where no information is contained in KDS predictions. We present these results in case 5. Table \ref{tab:concept_results} summarizes the results of all evaluated cases. We now further elaborate on results for each case

\begin{table*}[h]
\caption{Results obtained by KENN in comparison to those obtained by KDS and DNN when evaluated independently. Description column states different experimental setting used in each case. }
\label{tab:concept_results}
\vskip 0.1in
\centering
\begin{tabular}{|c|l|ccc|}
\hline
\multirow{2}{*}{Case} & \multicolumn{1}{c|}{\multirow{2}{*}{Description}} & \multicolumn{3}{c|}{MSE}                                              \\ \cline{3-5} 
                      & \multicolumn{1}{c|}{}                             & \multicolumn{1}{c|}{DNN}  & \multicolumn{1}{c|}{KDS}  & KENN          \\ \hline
1.                    & Full training data and accurate KDS               & \multicolumn{1}{c|}{7.15} & \multicolumn{1}{c|}{2.42} & \textbf{1.49} \\ \hline
\multirow{2}{*}{2}    & Reduced training data (50\%) and accurate KDS     & \multicolumn{1}{c|}{7.61} & \multicolumn{1}{c|}{2.42} & \textbf{1.54} \\ \cline{2-5} 
                      & Reduced training data (10\%) and accurate KDS     & \multicolumn{1}{c|}{8.01} & \multicolumn{1}{c|}{2.42} & \textbf{2.21} \\ \hline
3                     & Full training data and handicap KDS               & \multicolumn{1}{c|}{7.15} & \multicolumn{1}{c|}{8.26} & \textbf{4.41} \\ \hline
4                     & Reduced training data (10\%) and handicap KDS     & \multicolumn{1}{c|}{8.01} & \multicolumn{1}{c|}{8.26} & \textbf{6.38} \\ \hline
\multirow{2}{*}{5}    & Full training data and redundant KDS              & \multicolumn{1}{c|}{7.15} & \multicolumn{1}{c|}{9.04} & \textbf{7.16} \\ \cline{2-5} 
                      & Full training data and zero KDS predictions       & \multicolumn{1}{c|}{7.15} & \multicolumn{1}{c|}{611}  & \textbf{7.15} \\ \hline
\end{tabular}
\end{table*}

\begin{figure}[ht]
\label{fig:1}
\vskip 0.2in
\begin{center}
\centerline{\includegraphics[width=\columnwidth]{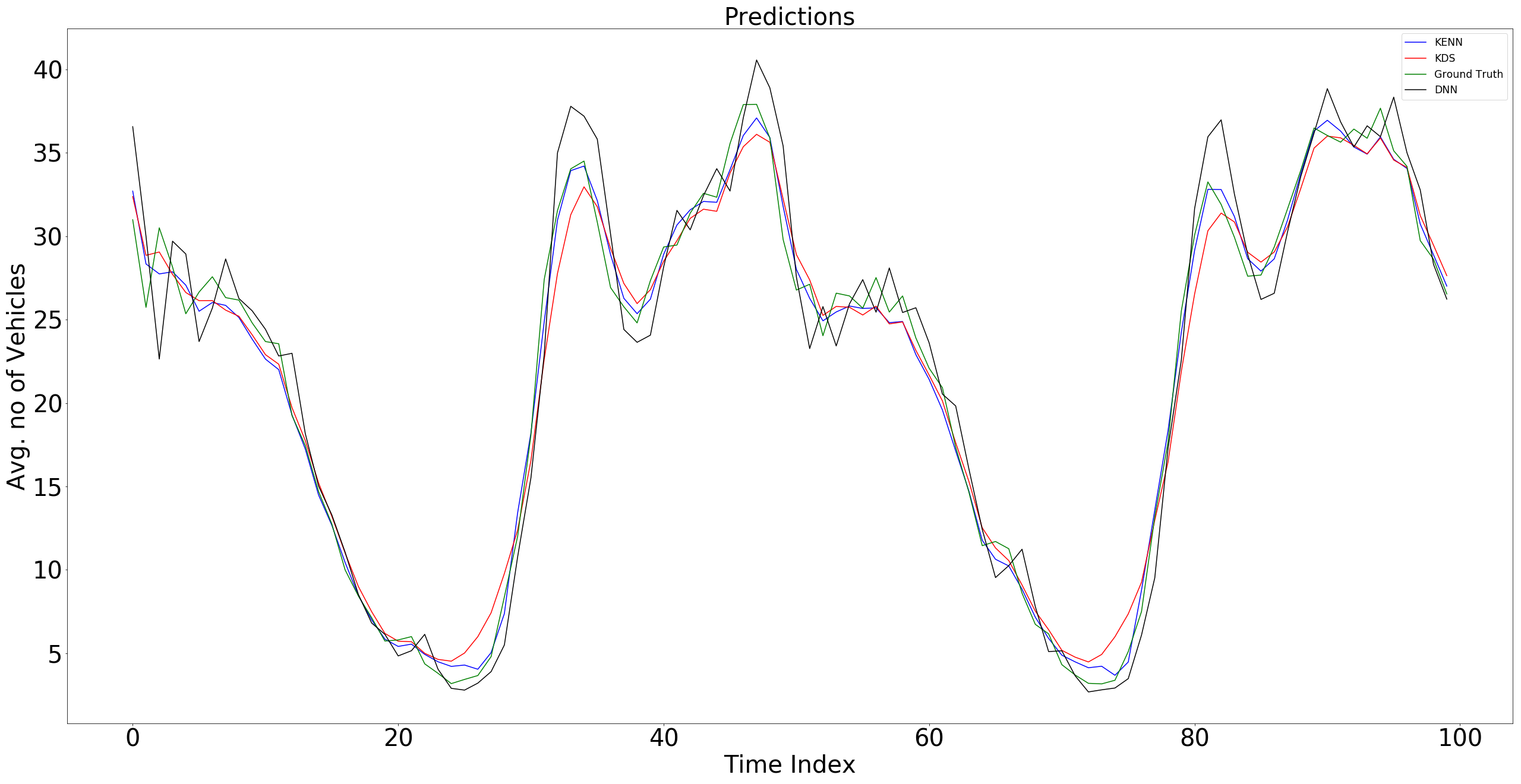}}
\caption{Predictions of first 100 observations of the test set for all the models.}
\label{icml-historical}
\end{center}
\vskip -0.2in
\end{figure}

\textbf{Case 1: Full training data and accurate KDS.} In this setting, we utilized the whole of the training data for training. We first evaluated MSE achieved by both KDS and DNN independently. KDS achieved a significantly better result when compared to the one achieved by DNN. This also highlights the fact that the time series domain is sometimes a challenge for DNNs, where simple methods often outperform complex DNN architectures. This phenomenon is also present in various other forecasting competitions~\cite{makridakis2000m3,makridakis2018m4}. However, the results achieved by KENN were significantly better than the results of both KDS and DNN. In particular, MSE achieved by KENN was almost \textbf{80\%} and \textbf{40\%} better than the loss achieved by DNN and KDS respectively. This highlights the proficiency of KENN in fusing information present in different domains. We visualize the results in Fig \ref{fig:1} where we plot the first 100 predicted observations in the test set for all the evaluated systems against the ground truth. It can be clearly seen that KENN dynamically adjusts the amount of information fused from the data domain, promoting predictions of KDS where they are accurate and making large corrections where KDS is less ideal. this phenomenon will be more profound in case 3, where we used handicapped KDS.  

\textbf{Case 2: Reduced Training data and accurate KDS.} Since one of our aims from the very beginning was to reduce the dependency of our framework on the data whilst keeping the strengths of data driven ML methods, we evaluate KENN in data scarce scenario. We achieve this by reducing the amount of training data available for training. We first reduce the training data by 50\% and evaluated all of our models. After which we took an extreme scenario where we reduced the training set by 90\% and utilized only 10\% of the training data. As KDS was not strictly dependent on the data, its performance was unaffected by the reduction in data. However, the performance of DNN did degrade.  Still, the prediction loss of KENN was significantly lower than both KDS and DNN for both scenarios. Perhaps one of the more surprising results was that with even only 10\% of the data, KENN was able to outperform DNN that was trained on the entire training set. 

\textbf{Case 3: Full training data and handicapped KDS.} In all of the previous scenarios, the KDS was comparatively better than the DNN employed in testing. Although the performance of KENN in previous experimental settings underscores KENN generalization capabilities where it worked equally well in data abundant and data scarce regimes, it is however imperative to show that this was not due to usage of an effective KDS system but was rather due to knowledge sharing capabilities of KENN. Moreover, it is also important to highlight that while reducing dependency on data we do not introduce an additional dependency on the accuracy of the KDS system. To demonstrate this we deliberately add random noise to predictions made by KDS effectively reducing their accuracy. The resulting MSE of this handicapped KDS is 8.26, which is significantly worse than the MSE of DNN. However, still KENN was able to extract useful information in both of the domains and produced results that were \textbf{43\%} better than the handicapped KDS and \textbf{34\%} better than the DNN architecture.
\begin{figure}[ht]
\label{fig:noisy}
\vskip 0.2in
\begin{center}
\centerline{\includegraphics[width=\columnwidth]{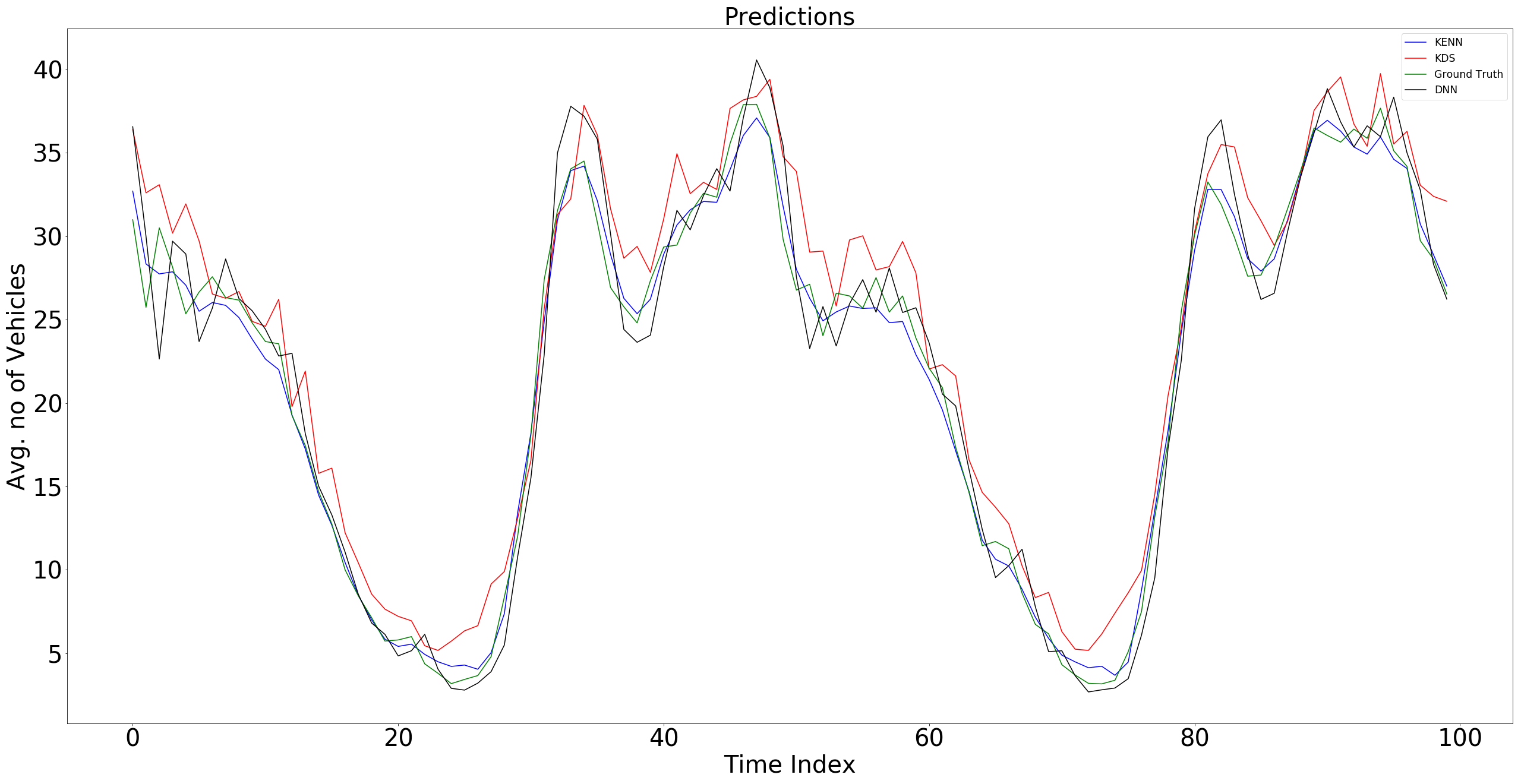}}
\caption{Predictions of first 100 observations of the test set for all the models. KDS predictions show the effect of random noise as they are quite in accurate}
\label{icml-historical}
\end{center}
\vskip -0.2in
\end{figure}

\textbf{Case 4: Reduced training data and handicapped KDS.} We also evaluated KENN in an extreme setting, where data is scarce and KDS is also handicapped. We believe that it is also important to analyze the performance of KENN when it is given a situation where weaknesses of both knowledge and data domains are combined. Here, we again reduce the training data by 90\% and used noisy KDS predictions.  However, despite these worst conditions KENN still managed to outperform both DNN and KDS and was still better than the DNN trained on 100\% of the training data. 

\textbf{Case 5: Full training data and redundant KDS.} Finally, to evaluate the performance of KENN in a scenario where no useful information is present in KDS we employ two settings. In the first setting, KDS was forced to repeat the last observed value of the time series as its prediction, essentially providing redundant information to KENN. In the second, KDS always gave zero value as its predictions. Hence there was no information for KENN. In both of these scenarios, KENN converges to the performance of DNN i.e the only source with useable information. This is a very strong result as it demonstrates that the loss of KENN is bounded and will always converge to the best-performing domain if the other domain lacks any useful information.

\section{Ablation Studies}\label{sec:Ablation}

One of the strengths of the knowledge domain is that knowledge is always expanding. With experience, our understanding of the problem increases, and as a result, so does our knowledge. Any fusion scheme should be fully capable of incorporating this changing knowledge otherwise it would not be able to fully utilize the strengths of the knowledge domain. Similarly, with the current rate of development ML methods are improving at a very fast pace. Hence, a fusion system must also be able to integrate new and improved DNN architectures. This is especially missing in most of the prior work~\cite{towell1994knowledge,tran2018deep} which are not very flexible in using different types of KDS or even DNN architecture. However, KENN is fully agnostic to the type of KDS or DNN architecture employed and as a result, it is fully capable of incorporating ever-changing knowledge and ML regimes. To demonstrate this we perform two additional experiments. In the first experiment, we completely change the KDS employed in KENN and employ a statistical method as our KDS namely Seasonal Autoregressive Integrated Moving Average (SARIMA). In the second setting, we change the DNN architecture, and instead of using LSTM based architecture, we employ a Temporal Convolutional Network (TCN). The results of both the experiments are shown in table \ref{tab:ab}

\begin{table}[h]

\caption{Results obtained when KENN when a different KDS and DNN were used in experiment 1 and 2 respectively}
\label{tab:ab}
\vskip 0.1in
\begin{center}
\begin{small}
\begin{sc}
\begin{tabular}{|l|ccc|}
\hline
\multicolumn{1}{|c|}{\multirow{2}{*}{Experiment Description}} & \multicolumn{3}{c|}{MSE}                                              \\ \cline{2-4} 
\multicolumn{1}{|c|}{}                             & \multicolumn{1}{c|}{DNN}  & \multicolumn{1}{c|}{KDS}  & KENN          \\ \hline
With SARIMA as KDS                                 & \multicolumn{1}{c|}{7.15} & \multicolumn{1}{c|}{2.56} & \textbf{1.51}  \\ \hline
With TCN as DNN                                    & \multicolumn{1}{c|}{7.01}   & \multicolumn{1}{c|}{2.42} & \textbf{1.52} \\ \hline
\end{tabular}

\end{sc}
\end{small}
\end{center}
\vskip -0.1in
\end{table}

As can be seen from table \ref{tab:ab}, the KENN framework is flexible and dynamic as it allows different KDS and DNN architecture to be integrated within the framework and does this while maintaining its core strengths of fusing relevant information from knowledge and data domains. This is very useful since it allows for changing rule sets within the KDS, where one can add, modify or remove clauses in the ruleset over time as the knowledge about the problem increases, without restructuring the complete architecture.

\section{Conclusion}

In this paper, we present a simple yet effective knowledge
fusion framework for time series forecasting, KENN, with
aim of combining strengths of knowledge and data driven
domains while suppressing their individual weaknesses. We
show that KENN successfully mitigates data dependency
of ML based methods and also works well even when the
knowledge contained in KDS is unreliable. This feat is
achieved due to the fusion of KDS in a residual connection which forces the ML base DNN to focus on finding
from the data domain only the information missing from
the knowledge domain. We show that KENN is agnostic
to the type of KDS and DNN architectures. We also show
that due to its less reliance on historical data, KENN also
beats SOTA methods across various time series datasets.
We believe that having a framework that makes use of both
knowledge and data driven methods in a complementary
manner is imperative towards unlocking the true potential of
AI. We think that KENN will not only improve the applicability
of AI methods in various real-world problem but will also
enable small scale industries, that do not have a lot of data,
to compete with giants of the AI industry.

\nocite{langley00}

\bibliography{KENN_main}
\bibliographystyle{icml2022}

\newpage
\appendix
\onecolumn
\section{Results Reproducibility }
We provide all the code to reproduce the results along with the optimal weights in the supplementary material. however we could not upload the data or the weights of the model due to their size. we will upon acceptance make the github public 

\section{Formulation of KDS}

\begin{figure}[ht]\label{fig:graph}
\vskip 0.2in
\begin{center}
\centerline{\includegraphics[width=\columnwidth]{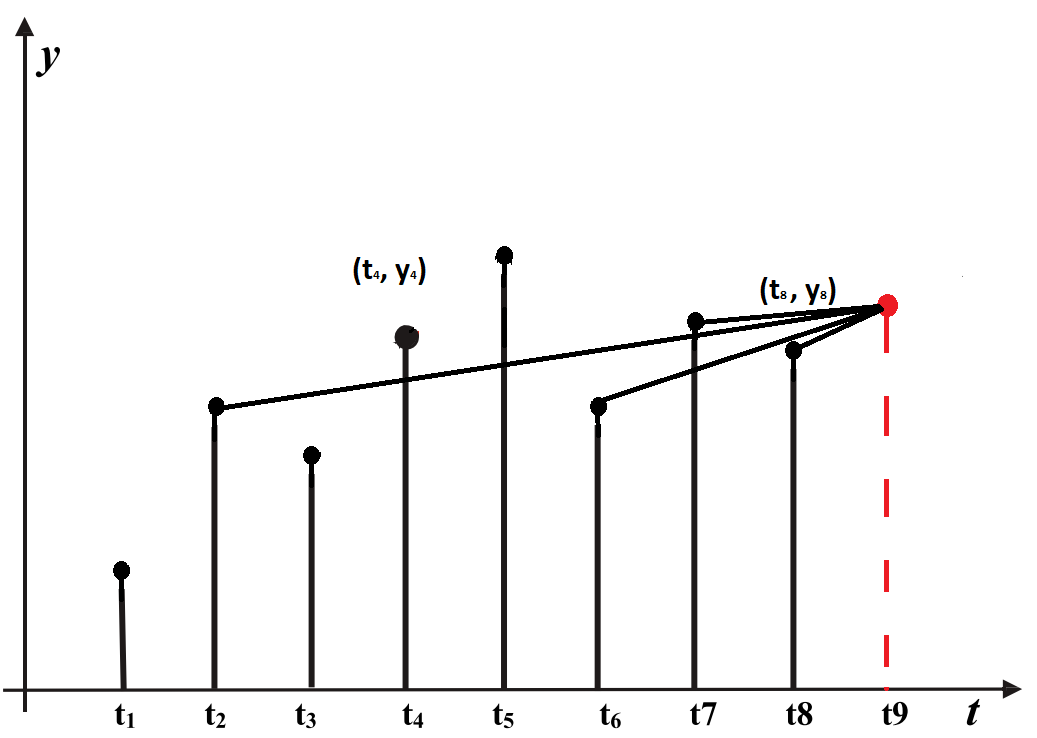}}
\caption{This figure illustrates conversion of time series into a graph consisting of nodes. Each observed value is represented by the height of node and the horizontal axis shows the time index. The nodes that have high similarity are connected via the edges.  }
\label{icml-historical}
\end{center}
\vskip -0.2in
\end{figure}

For our knowledge base in KDS, we use concepts drawn from graph theory. Specifically we consider each observed value at time step $t$ as a node. In graph theory, nodes are interconnected where the connection between two nodes is based on some similarity measure. In our case we use Partial auto-correlation function (Pacf) as our similarity measure. First pacf is calculated for the training set. Only those nodes are connected that have a correlation value of $>0.2$. This is shown in Figure \ref{fig:graph} and \ref{fig:pacg} where the resulting graph and corresponding pacf values are shown. These figures are only for illustration and do not represent any real data.

\begin{figure}[ht]\label{fig:pacf}
\vskip 0.2in
\begin{center}
\centerline{\includegraphics[width=\columnwidth]{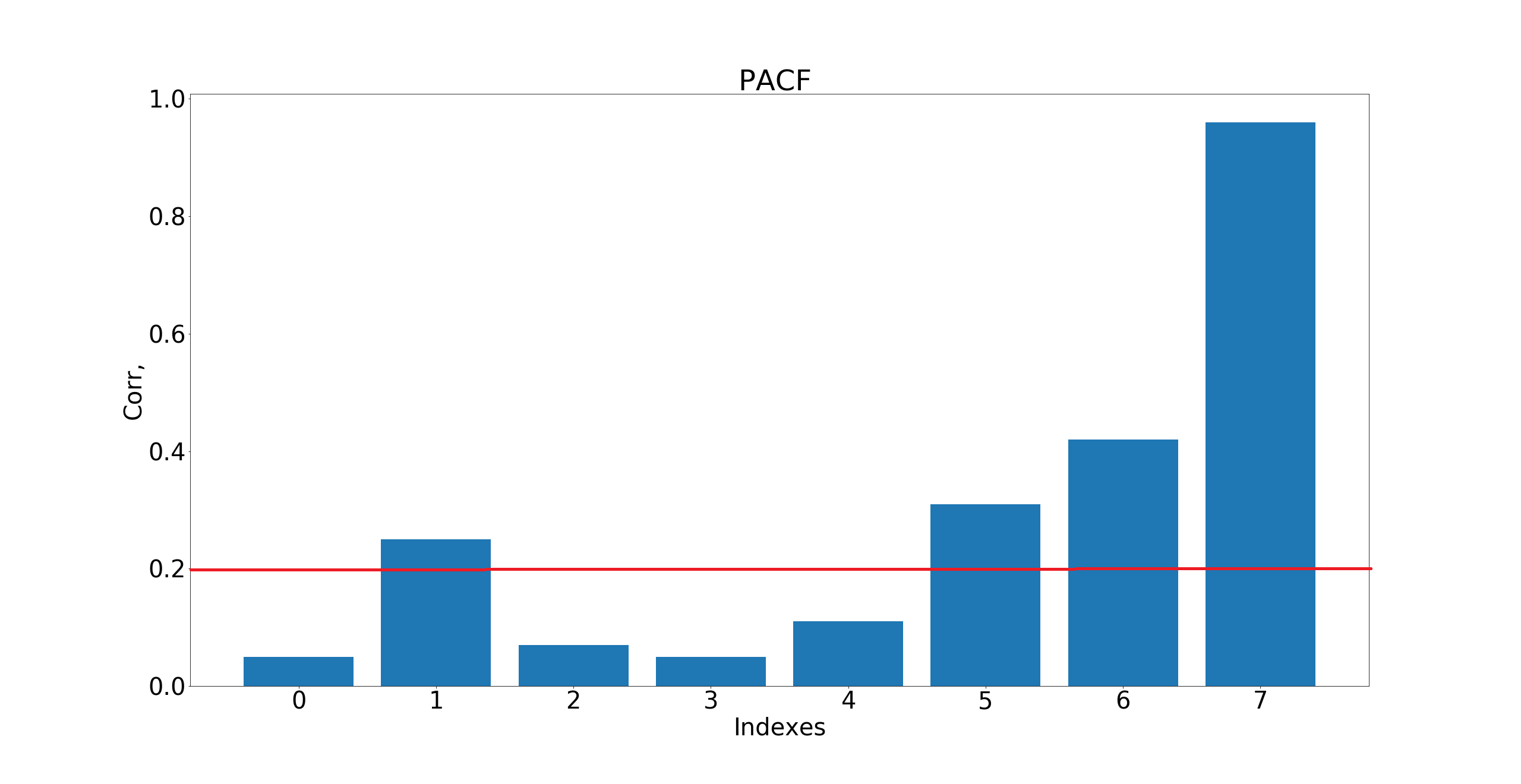}}
\caption{This figure show the values of Pacf function. The cut off threshold is set at 0.2 and only those nodes are connected whose pacf $>$ 0.2 as illustrated in Fig. 1 of supplementary materials.  }
\label{icml-historical}
\end{center}
\vskip -0.2in
\end{figure}

The edges have a associated edge weight that shows the overall strength of the connection. The edge weights $Ew$ between two connected node $n_1$ and $n_2$ are calculated by the following equation:
\begin{equation}
    {Ew}_{n_{1},n_{2}} = pacf(n_{2}) * \delta 
\end{equation}
\begin{equation*}
    \text{      where }  \delta=\begin{cases} 1 & \text{if } (t_{n_{1}}-t_{n_{2}})\leq 48\\ \frac{1}{2\log_2(t_{n_{1}}-t_{n_{2}})} &\text{otherwise} \end{cases}
\end{equation*}

The edge weights are penalized according to distance between them. If they are far apart in time their weights are reduced by factor $\delta$. Since data represents average traffic flow per 30 mins, 48 represents one day. Hence as soon as the temporal distance between the connected nodes becomes greater than one day the penalization comes into effect. The forecast at time $t_9$ is then calculated by the following equation
\begin{equation}
      \hat{\mathcal{Y}}_{KDS} = \sum_{x=1}^{n} {{Ew}_{n_{9},x}*Y_{t_x}} \text{  ,where} \sum{Ew} =1
\end{equation}

where $n$ defines the total number of connected nodes to $n_9$ and $Y_{t_x}$ represents observed value of time series at time $t_x$. Another set of rules are then applied on $\hat{\mathcal{Y}}_{KDS}$ that checks mean of the same 2 hours of the previous day and tires to match the mean if the variation is within 0.7 standard deviation. Basically modelling the fact that there is high seasonality effect during peak hours etc.

\end{document}